\pdfoutput=1

\documentclass[11pt]{article}
\usepackage{comment}
\usepackage{multirow}
\usepackage{naacl2021}
\usepackage{times}
\usepackage{latexsym}
\usepackage{graphicx}
\graphicspath{ {./Images/} }

\usepackage[T1]{fontenc}

\usepackage[utf8]{inputenc}

\usepackage{microtype}

\title{Pre-Training a Language Model Without Human Language}

\author{Cheng-Han Chiang \\
  National Taiwan University, Taiwan \\
  \texttt{dcml0714@gmail.com} \\\And
  Hung-yi Lee \\
  National Taiwan University, Taiwan \\
  \texttt{hungyilee@ntu.edu.tw} \\}

\begin{document}
\maketitle
\begin{abstract}
In this paper, we study how the intrinsic nature of pre-training data contributes to the fine-tuned downstream performance.
To this end, we pre-train different transformer-based masked language models on several corpora with certain features, and we fine-tune those language models on GLUE benchmarks. 
We find that models pre-trained on unstructured data beat those trained directly from scratch on downstream tasks.
Our results also show that pre-training on structured data does not always make the model acquire ability that can be transferred to natural language downstream tasks.
To our great astonishment, we uncover that pre-training on certain non-human language data gives GLUE performance close to performance pre-trained on another non-English language.
\end{abstract}

\section{Introduction}
Neural language models (LMs) are prevalent in nowadays natural language processing (NLP) community, and they are indispensable to a variety of NLP tasks.
Researchers have devoted themselves to understanding what these models have learned and how they work. 
Probing a trained model is widely used to understand to what extent a model learns certain linguistic features~\citep{kovaleva2019revealing, hewitt2019structural, tenney2019bert, tenney2018you, lin2019open}.
Another line of research focuses more on how training corpora affect the trained LMs ~\citep{micheli2020importance, gururangan-etal-2020-dont, zhang2020you}. 

\begin{figure*}[h]
\centering
\includegraphics[clip, trim = 0px 195px 250px 0px,width=0.75\linewidth]{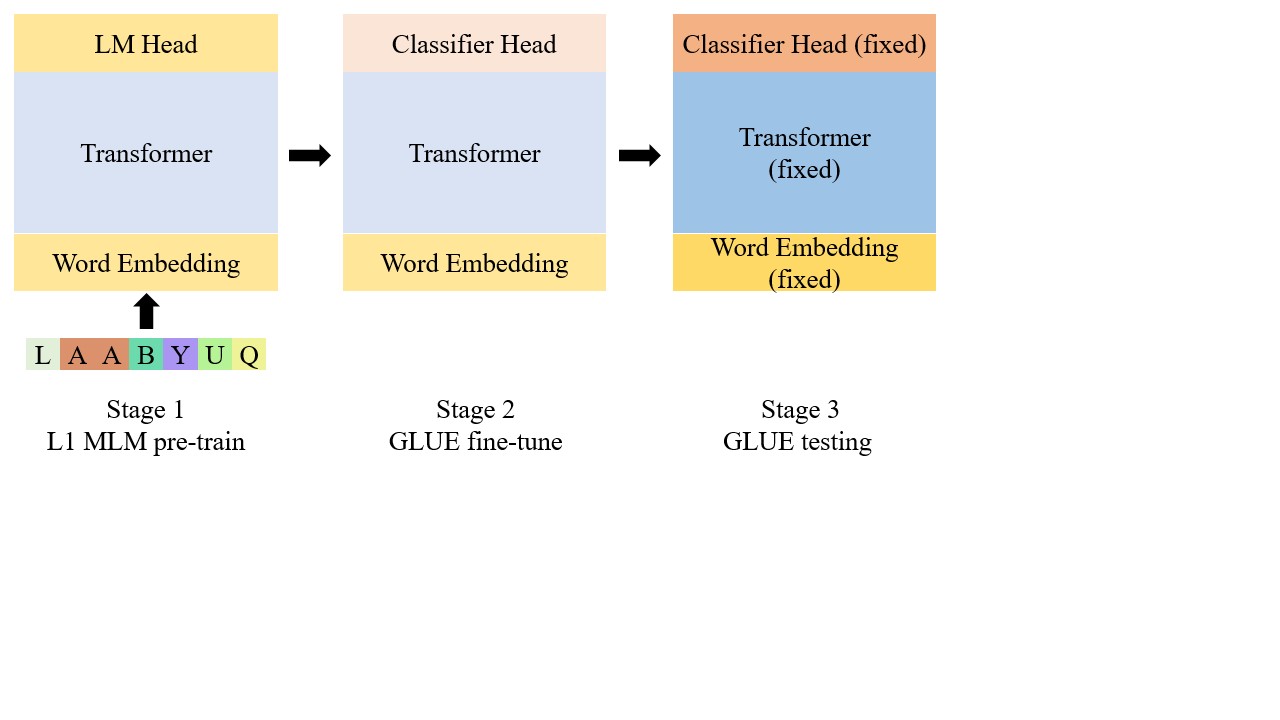}
\caption{Work flow of our experiments: We first pre-train the whole masked language model on L1 (protein sequence in this figure), and fine-tune the whole model on English downstream tasks. We then test the performance on the fine-tuned downstream task.
It takes about 3 days to finish the whole process on a single V100.}
\label{fig:exp}
\end{figure*}

In this work, we aim to understand how downstream performance varies across models pre-trained on data of particular traits.
The core problem we determine to answer is: What factors in the pre-training data make a pre-trained transformer LM perform better on downstream tasks than their trained from scratch counterparts?
To answer this question, we pre-train many different transformer LMs on dataset from miscellaneous disciplines, ranging from amino acid sequences in complex living organisms to artificial data generated by a simple python script.
We then fine-tune them on English downstream tasks.
The process is illustrated in Figure~\ref{fig:exp}. 

Recently,~\citet{papadimitriou2020learning} proposed to train an LSTM LM on a non-natural language dataset and test the LM's perplexity on natural language. 
They observed that LSTM LM trained on structured dataset gives perplexity far lower than those trained on unstructured data.
While the observations are intriguing, this setting doesn't match the common setting widely applied nowadays, in which we fine-tune pre-trained LMs on downstream tasks.
This is the first paper investigating whether masked language model (MLM) pre-training on non-natural language aids downstream natural language tasks' performance.

Based on the experiments, we have the following observations: 
\begin{itemize}
\item We reveal that fine-tuning models pre-trained on unstructured data outperforms model trained from scratch on downstream tasks. 
\item We find that structured pre-training data is not a sufficient condition to a pre-trained model that can perform well on NLP tasks.
\item We discover that pre-training on a simple artificial dataset with hierarchical structure leads to downstream performance comparable to models pre-trained on human language.
\item Our experiments show that token distribution is not the key factors to how well the model transferred to downstream tasks,  while the number of token embeddings used during pre-training affects downstream performance.
\end{itemize}

\section{Experiment Setups}
\label{sec:setup}
In our experiments, we pre-train \emph{n} RoBERTa-base~\citep{liu2019roberta} models on \emph{n} different types of pre-training data.
We call the pre-training data L1 (first language).
We then evaluate the pre-trained models' ability by fine-tuning them on different downstream tasks.
The overall workflow is illustrated in Figure~\ref{fig:exp}.
We adopt the classic GLUE ~\citep{wang2019glue} benchmarks to evaluate the models pre-trained on different L1s while excluding WNLI following ~\citet{devlin2019bert}.
For each task, we use a certain set of hyperparameters and the same random seed to fine-tune the model, and we report the results on the evaluation set. 
Details regarding all experiments can be found in Appendix~\ref{app: pretrain}.

Our experiment setup may seem to resemble the Test for Inductive Bias via Language Model Transfer (TILT) proposed in ~\citet{papadimitriou2020learning} at first sight, which pre-trains an LSTM LM on L1, follows by only fine-tuning word embeddings on Spanish, and test the perplexity on Spanish.
However, the main purpose of TILT is to analyze the encoding of grammatical structure in LMs, so they do not fine-tune LSTM on Spanish.
On the contrary, our goal is to understand what factors in pre-training data make the pre-trained model perform better than models trained from scratch on downstream tasks.

\section{Pre-training Data}
\label{subsubsec: artificial}
We use two baseline pre-training dataset for our experiments: the \textbf{random baseline} and the \textbf{Zipf baseline}, both corpora have 29995 tokens, excluding 5 special tokens.
For the random baseline, we draw the tokens from a uniform distribution and form sequences with a length of 90 to 120 tokens.
For the Zipf baseline, we sample the tokens from the same uni-gram distribution of English.
We also pre-train an \textbf{English} MLM with a subset of the English Wikipedia to serve as the performance upper bound.
The pre-training corpora size is around 80MB for the previous three datasets.

We select several pre-training corpora in distinct disciplines that contain structure, including a biological dataset, a programming language corpus, an artificial dataset with a hierarchical structure, and a human language.

The biological dataset we adopt is \textbf{amino acid} sequence corpora obtained from ~\citet{min2019pre}.
The characteristic of a protein is determined by its primary structure, i.e. the amino acid sequence. 
Chemical bonds between amino acids determine the secondary and tertiary structure of the folded protein, which further determines the functions of the protein.
We use the one-letter abbreviation (A-Z) to represent each amino acid, and the total number of tokens in this dataset is 36M.

For programming language, we use Habeas corpus from ~\citet{movshovitz2013natural}, which contains tokenized \textbf{Java script}. 
We use the code from ~\citet{papadimitriou2020learning} to extract the data and remove tokens that are labeled as a comment, making the training corpus contain only programming language.
The total number of tokens in the pre-training data is 10M, and the vocabulary size of the model is 30K.

\begin{figure}[h]
\centering
\includegraphics[clip, trim = 30px 195px 30px 195px,width=1.0\linewidth]{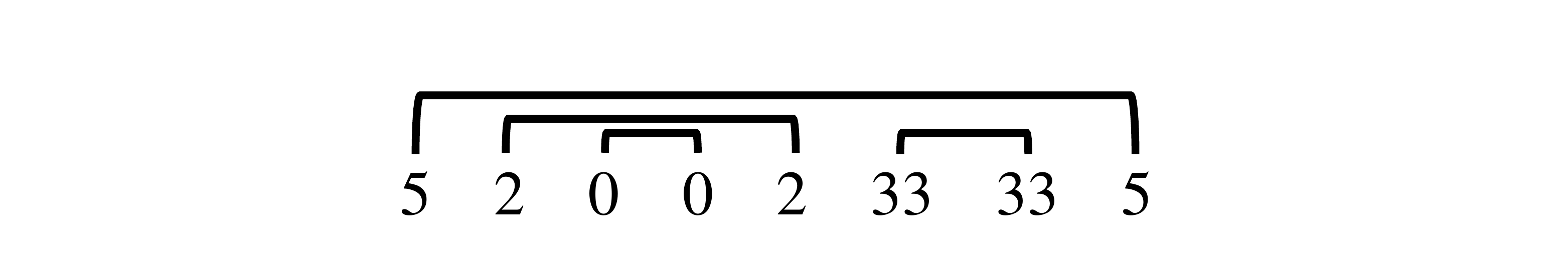}
\caption{An illustration of the artificial dataset.}
\label{fig:artificial}
\end{figure}


The \textbf{artificial} dataset we construct has a vocabulary size of 28996, and the total number of tokens in training data is 23.5M.
The dataset is generated by the following stack-based grammar, following ~\citet{papadimitriou2020learning}: At each time step \emph{t}, we sample \(X_t\) from a Bernoulli distribution with \(P(X_t = 1) = 0.4\).
If \(X_t=1\), we sample a token based on English's uni-gram distribution, place the sampled token at position \(t\) of the generated sequence, and push the same token into the stack.
When \(X_t=0\), we pop the top element of the stack and put the popped token at position \(t\) of the generated sequence.
Figure~\ref{fig:artificial} shows a simple example.
We can observe from Figure~\ref{fig:artificial} that sequence generated in this manner contains a nesting hierarchical parentheses structure, which is similar to the dependency tree structure in natural language.

The last dataset used is a human language.
We select a human language different from downstream tasks to compare the effect of non-human language pre-training data.
We use \textbf{Kannada} from OSCAR dataset ~\citep{suarez2020monolingual}.
Kannada is a language predominantly spoken by the people in the southern western region of India.
The main reason we choose this dataset lies in its subject(S)-object(O)-verb(V) structure, different from the S-V-O structure of our target language used in fine-tuning.
The pre-training corpora size is 160MB, and the vocabulary size used in pre-training is 30K.

\begin{table*}[h!]
\centering
\begin{tabular}{ccccccccccc}
\hline
&\textbf{L1} & \textbf{STS-B} & \textbf{QNLI} & \textbf{QQP} & \textbf{CoLA} & \textbf{SST-2} & 
\textbf{MNLI} & \textbf{MRPC} & \textbf{RTE} & \textbf{Avg}\\
\hline
\multirow{2}{*}{1} & \textbf{No Pre-train}  & 0.17 & 0.60 & 0.75 & 0.13 & 0.83  & 0.64 & 0.67 & 0.50 & 0.54\\
&\textbf{Pre-train En} & 0.76 & 0.83 & 0.86 & 0.34 & 0.88 & 0.76 & 0.77 & 0.53 & 0.72\\
\hline
\multirow{2}{*}{2} & \textbf{Rand. Baseline} & 0.29 & 0.66 & 0.80 & 0.14 & 0.83 & 0.65 & 0.77 & 0.51 & 0.58\\
&\textbf{Zipf Baseline} & 0.38 & 0.66 & 0.80 & 0.11 & 0.82 & 0.64 & 0.81  & 0.49 & 0.59\\
\hline
\multirow{3}{*}{3} & \textbf{Amino Acid} & 0.16 & 0.65 & 0.79 & 0.07 & 0.82 & 0.60 & 0.81 & 0.44 & 0.55\\
&\textbf{Java Script} & 0.31 & 0.67 & 0.77 & 0.02 & 0.81 & 0.66 & 0.75 & 0.51 & 0.56\\
&\textbf{Kannada} & 0.76 & 0.77 & 0.83 & 0.12 & 0.81 & 0.69 & 0.80 & 0.55 & 0.67\\
\hline
\multirow{2}{*}{4} & \textbf{Artificial (Uni.)} & 0.72 & 0.77 & 0.80 & 0.14 & 0.81 & 0.69 & 0.77 & 0.52 & 0.65\\
&\textbf{Artificial (Zipf)} & 0.76 & 0.77 & 0.83 & 0.11 & 0.82 & 0.69 & 0.75 & 0.53 & 0.66\\
\hline
\multirow{4}{*}{5} & \textbf{Artificial (5000)} & 0.73 & 0.76 & 0.82 & 0.09 & 0.84 & 0.69 & 0.82 & 0.53 & 0.66\\
&\textbf{Artificial (500)} & 0.42 & 0.68 & 0.80 & 0.08 & 0.81 & 0.68 & 0.79 & 0.51 & 0.60\\
&\textbf{Artificial (50)} & 0.18 & 0.62 & 0.74 & 0.06 & 0.82 & 0.61 & 0.77 & 0.52 & 0.54\\
&\textbf{Artificial (50-s)} & 0.65& 0.73 & 0.84 & 0.06 & 0.80 & 0.64 & 0.75 & 0.50 & 0.62\\
\hline
\end{tabular}
\caption{Downstream results of different pre-trained models, and the model trained from scratch on downstream tasks (no pre-train in the first row). 
The evaluation metrics of MRPC and QQP are F1 score, Spearman correlation coefficient is reported for STS-B, and the rest tasks are evaluated with accuracy.
Results of MNLI are the average of matched and mismatched.
Please refer to Section~\ref{sec: distribution} and Section~\ref{sec: number of tokens} for the meaning of parentheses in the last two blocks.
50-s stands for 50-substitute in Section~\ref{sec: number of tokens}.
Abbreviation used: En: English, Rand: random, Uni: uniform.}
\label{tab:structure}
\end{table*}

\section{Experiments and Results}
The overall results are illustrated in Table~\ref{tab:structure}.
In this section, we discuss how certain aspects of the pre-training corpora affect how \emph{good} a model can become.
By the word \emph{good}, we refer to the model's ability to be fine-tuned on downstream tasks, which is the performance improvement over training the model from scratch on downstream tasks.


\subsection{Is Structured Data All You Need For Pre-training?}
\label{sec: structure}
We intend to answer this question: Is structured data the key to a \emph{good} pre-trained model?
We compare the models pre-trained on structured data with models pre-trained on unstructured baselines. 
If the downstream performance of models pre-trained on structured data can beat their unstructured counterparts, then we may conclude that structure in the pre-training data is a key factor in the success of pre-trained transformer language models.

From the first two blocks of Table~\ref{tab:structure}, we find that models pre-trained on unstructured data outperform the models trained from scratch.
This suggests that the pre-trained model can still aid downstream performance, albeit the seemingly meaningless pre-training corpora.

From the third block in Table~\ref{tab:structure}, we find that pre-training on structured data may not always lead to a better model.
Models pre-trained on amino acid and Java scripts are almost on a par with the models trained from scratch.
Not much to our surprise, the model pre-trained on Kannada performs far better than the two baseline models.

Amazingly, fine-tuning the model pre-trained on artificial data gives comparable performance compared with the model pre-trained on Kannada.
This implies that it might be worth trying to pre-train a model on this kind of hierarchical nesting structured dataset, and fine-tune the model on some low resource languages to obtain decent downstream performance.
The artificial dataset consists of no semantic knowledge useful for downstream natural language tasks, so it is reasonable to infer that most knowledge the model learns from pre-training is the skill to model the hierarchical structure and long-term dependency.
Equipped with this ability, the model can outrun models trained from unstructured data.

Our results show that models benefit from pre-training on a certain type of structured corpora, while not every structured corpus leads to a good pre-trained model for NLP downstream tasks.



\subsection{Does Pre-training Data Token Distribution Affect the Performance on Downstream Tasks?}
\label{sec: distribution}

We notice that the two baseline models' performance is similar in almost all downstream tasks. 
This indicates that the uni-gram distribution of tokens in the training corpora makes little difference to the downstream performance when the pre-training data themselves are unstructured. 
We further ask whether this is also the case when the data is structured.
We construct the artificial dataset as in Section~\ref{subsubsec: artificial}, and aside from sampling based on Zipf distribution, we create another dataset whose tokens are sampled from the uniform distribution over tokens except for special tokens.
The results, demonstrated in the fourth block in Table~\ref{tab:structure}, show that even when the pre-training data is structured, token distribution still has little influence on how well the model can be fine-tuned.


\subsection{Does Token Numbers Mismatch between Pre-training and Fine-tuning Affect Downstream Performance?}
\label{sec: number of tokens}
This section investigates whether the mismatch between vocabulary size during pre-training\footnote{The number of different tokens in pre-training data.} and fine-tuning contributes to how well the pre-trained model performs on downstream tasks.
To study the influence of vocabulary size, we construct different artificial data by sampling tokens from different bin sizes (50, 500, and 5000).
While the vocabulary size during pre-training is different for those models, their actual word embedding table sizes are still the same.

From the last block in Table~\ref{tab:structure}, we observe that the averaged performance significantly degrades in the case when only 50 tokens are used during pre-training, while the performance gradually recover when the token number mismatch between pre-training and fine-tuning narrows.
Tokens appearing in the pre-training data receive disproportionately larger gradients than tokens not in the pre-training data during pre-training, and this artifact cripples the downstream performance.

The above observation make it hard to tell whether model pre-trained with amino acid sequence failed to perform well on downstream tasks due to the token number mismatch.
Thus, we conduct further experiments to remove the undesirable artifact arise from the mismatch.
Say we only use the first 50 tokens (excluding special tokens) during pre-training while the rest 29950 token embeddings are not used, then before fine-tuning the model on downstream tasks, we substitute those unused token embeddings with those 50 used token embeddings.
We call the above setting 50-substitute.
In this case, different tokens will share the same token embeddings when the model starts to be fine-tuned. 

From the last row in Table~\ref{tab:structure}, we find that the model recovers its ability to be fine-tuned when pre-trained on artificial dataset.
However, when performing the same substitution on the model pre-trained with amino acid, the model still fail to be fine-tuned.
Together with Section~\ref{sec: structure}, we can conclude that the main reason a pre-trained model failed to transfer to human language downstream tasks lies in the intrinsic property of the pre-training data.

\subsection{Further Fine-tuning with English MLM before Fine-tuning on GLUE}
\label{subsec: MLM finetune}
It is innate to fine-tune the word embeddings of pre-trained models on English before fine-tuning on GLUE.
This is for aligning the word embeddings of L1 acquired during pre-training with the word embeddings of English.
We conduct experiments similar to Table~\ref{tab:structure}, and the only difference lies in that we fine-tune the word embeddings and language model head of the pre-trained model with MLM on English before fine-tuning on GLUE. 
We find the performance slightly advance mostly, with improvement in Java script being the most salient.
We leave detailed results in Appendix~\ref{app: finetune}.

\section{Conclusion}
We study how pre-trained data might and might not affect the downstream performance of a transformer-based pre-trained LM. 
We find that fine-tuning with models pre-trained on data without any structures can surpass performance obtained directly trained from scratch on downstream tasks.
Our results also show that pre-training with structured non-human language corpora does not always equip the model to perform competently on downstream tasks in general. 
We also discover that pre-training on a certain artificial dataset gives downstream performance comparable to pre-training on another natural language.
We reveal that token distribution in the pre-training corpora merely affects pre-trained model performance on downstream tasks.
Last, our experiments show that the number of token embeddings used during pre-training greatly contribute the downstream performance, while this can be mitigate by some manipulations on the token embeddings in certain cases.
We hope our analysis provides insights into what kind of pre-training data makes a pre-trained model a pre-trained model.



\section*{Broader Impact}
We find an surprising simple artificial dataset to pre-train an language model, and we believe that our work have the potential to be applied to low-resource language when pre-training data are scarce.
We think our work do not cause any ethical issues.


\bibliography{anthology,custom}
\bibliographystyle{acl_natbib}
\clearpage
\appendix
\section{Experiment Details}
\label{app: pretrain}
We give detailed model architectures of our RoBERTa-base model and hyperparameters used in pre-training.
\subsection{Model}
We use RoBERTa-base, a 12-layered transformer model with hidden dimension 768 and 12 attention heads per layer. 
The total number of parameters of the model is around 110M.
We pre-train RoBERTa using Huggingface~\citep{Wolf2019HuggingFacesTS} code base.
\subsection{Hyperparameters}
The hyperparameters used in all pre-training experiments are listed in Table~\ref{tab:pretrain hyperparam}
\begin{table}[h]
    \centering
    \begin{tabular}{|c|c|}
    \hline
        Batch size & 150 \\
        Learning rate & 5E-5 \\
        Total steps & 200K \\
        Warmup steps & 10k\\
        Max Position & 128\\
    \hline
    \end{tabular}
    \caption{Pre-training hyperparemeters for BERT.}
    \label{tab:pretrain hyperparam}
\end{table}
\subsection{Pre-training Data}
We put all details related to all pre-training data in Table~\ref{tab:pretrain data}.
We provide download link to the pre-training dataset, along with the training and validation loss at the end of pre-training.
The artificial data and baseline dataset can be generated following the script in our code.
The train/evaluation split can be found in the supplementary materials.
We also include the vocabulary size (including special tokens) of each model on the last column.
The vocabulary file is obtained by training a WordPiece tokenizer on the training data for Java, Kannada, and Wikipedia dataset.
\begin{table*}[h]
    \centering
    \begin{tabular}{ccccc}
    \hline
        \textbf{Dataset} & \textbf{Link} & \textbf{Training Loss} & \textbf{Eval. Loss} & \textbf{Vocab Size}\\
        \hline
        Wikipedia & \href{https://dumps.wikimedia.org/enwiki/latest/enwiki-latest-pages-articles.xml.bz2}
        {Wikidump}&2.204&3.354&30000\\
        Java &  \href{https://github.com/habeascorpus/habeascorpus-data-withComments}{Java data} &0.03227&1.025&30000\\
        Amino Acid & \href{http://ailab.snu.ac.kr/PLUS/data_compressed/Pfam.tar.gz}{PLUS} &2.041&2.201&28895\\
        Kannada & \href{https://oscar-corpus.com/}{OSCAR} &2.366&3.128&30000\\
        Random baseline & NA &9.428&9.467& 30000\\
        Zipf Baseline & NA &6.351&6.446& 30000\\
        Artificial (Uniform) & NA &1.996&2.409& 29991\\
        Artificial (Zipf) & NA &1.599&1.774& 29991\\
        Artificial (50) & NA &1.558&1.754& 29991\\
        Artificial (500) & NA &1.563&1.762& 29991\\
        Artificial (5000) & NA &1.548&1.701& 29991\\
    \hline
    \end{tabular}
    \caption{Details for dataset used in pre-training.}
    \label{tab:pretrain data}
\end{table*}
\subsection{Fine-tuning Details}
We fine-tune GLUE using Huggingface~\citep{Wolf2019HuggingFacesTS} code base.
The model fine-tuned in this section is RoBERTa base with classifier on top of the last transformer layer.
The whole model fine-tuned is has 110M parameters.
\subsubsection{Dataset}
We provide statistics on the 8 GLUE tasks we used in Table~\ref{tab:downstream_st}
\begin{table}[]
    \centering
    \begin{tabular}{c|c}
       Task  & Examples \\
       \hline
       MRPC & 3.6K / 0.4K / 1.7K \\
       RTE & 2.4K / 0.2K / 3K \\
       STS-B & 5.7K / 1.5K / 1.3K \\
       QNLI & 104K / 5.4K / 5.4K \\
       QQP & 363K / 40.4K / 391.0K\\
       CoLA & 8.5K / 1.0K / 1.1K \\
       MNLI & 392.7K / 9.8K + 9.8K / 9.8K + 9.8K\\
       SST-2 & 67.4K / 0.9K / 1.8K\\
    \end{tabular}
    \caption{Statistics of (train / dev/ test) in GLUE tasks
    MNLI contains matched and mismatched in dev and test set. We didn't evaluate our models' performance on test set.}
    \label{tab:downstream_st}
\end{table}
\subsubsection{Fine-tuning Hyperparameters}
We list the hyperparameters used in fine-tuning GLUE in Tabel~\ref{tab:glue_hpp}.
\begin{table*}[ht]
    \centering
    \begin{tabular}{c|cccccccc}
    & LR & BSZ & RoBERTa DR & Classifier DR & TS & WS & MSL \\
    \hline
    CoLA &1.00E-05& 16 &0 &0.1&5336 &320 &128 \\
    STS-B &2.00E-05& 16 &0 &0.1 &3598& 214 &128\\
    SST-2& 1.00E-05 &32& 0& 0.1 &20935 &1256 &128 \\
MNLI& 3.00E-05 &128& 0 &0.1 &10000 &1000& 128\\
QNLI& 1.00E-05 &32& 0 &0.1& 33112& 1986& 128\\
QQP& 5.00E-05& 128& 0 &0.1& 14000 &1000 &128\\
RTE &3.00E-05& 32& 0 &0.1 &800 &200 &128\\
MRPC& 2.00E-05& 32& 0 &0.1& 800& 200 &128\\
SQuAD2.0 &3.00E-05& 48& 0 &0.1& 8144& 814 &128
    \end{tabular}
    \caption{Hyperparameters for ALBERT in downstream tasks. LR: Learning Rate. BSZ: Batch Size. DR: Dropout Rate. TS: Training Steps. WS: Warmup Steps. MSL: Maximum Sequence
Length}
    \label{tab:glue_hpp}
\end{table*}

\subsection{Resource}
Out computation resource is V100 GPU.
Pre-training a RoBERTa following our parameters given in ~\ref{tab:pretrain hyperparam} takes 60 hours on a single V100, and fine-tuning the pre-trained models on the 8 GLUE tasks following hyperparameters in ~\ref{tab:glue_hpp} takes about 12 hours on a V100.

\section{Fine-tune the Model on English MLM Before Fine-tuning on GLUE}
\label{app: finetune}
This is the detailed experiment data for Section~\ref{subsec: MLM finetune}
\begin{table*}[h!]
\centering
\begin{tabular}{cccccccccc}
\hline
\textbf{L1} & \textbf{STS-B} & \textbf{QNLI} & \textbf{QQP} & \textbf{CoLA} & \textbf{SST-2} & 
\textbf{MNLI} & \textbf{MRPC} & \textbf{RTE} & \textbf{Avg}\\
\hline
\textbf{No Pre-train}  & 0.17 & 0.60 & 0.75 & 0.13 & 0.83  & 0.65 & 0.67 & 0.50 & 0.54\\
\textbf{Pre-train on English} & 0.76 & 0.83 & 0.86 & 0.34 & 0.88 & 0.76 & 0.77 & 0.53 & 0.72\\
\hline
\textbf{Random Baseline} & 0.28 & 0.67 & 0.80 & 0.12 & 0.83 & 0.66 & 0.71 & 0.56 & 0.57\\
\textbf{Zipf Baseline} & 0.34 & 0.71 & 0.81 & 0.17 & 0.84 &0.67 & 0.81  & 0.53  & 0.61\\
\hline
\textbf{Amino Acid} & 0.24 & 0.65 & 0.79 & 0.07 & 0.82 & 0.65 & 0.75 & 0.50 & 0.56\\
\textbf{Java Script} & 0.25 & 0.78 & 0.82 & 0.12 & 0.82 & 0.71 & 0.78 & 0.51& 0.60\\
\textbf{Kannada} & 0.79 & 0.78 & 0.84 & 0.15 & 0.85 & 0.71 & 0.81 & 0.57 & 0.69\\
\hline
\textbf{Artificial (Uniform)} & 0.73 & 0.79 & 0.82 & 0.17 & 0.82 & 0.71 & 0.75 & 0.55 & 0.67\\
\textbf{Artificial (Zipf)} & 0.79 & 0.79 & 0.83 & 0.11 & 0.82 & 0.72 & 0.75 & 0.57 & 0.67\\
\hline
\end{tabular}
\caption{Downstream results of different pre-trained models, and the model trained from scratch on downstream tasks (no pre-train in the first row). 
The evaluation metric of MRPC and QQP are F1 score, spearman correlation coefficient is reported for STS-B, and the rest tasks are evaluated with accuracy. 
Result of MNLI is averaged between matched and mismatched.
Please refer to Section~\ref{sec: distribution} and Section~\ref{sec: number of tokens} for the meaning of parentheses in the last two blocks.}
\label{tab:structure finetune}
\end{table*}
\begin{figure*}[ht]

\centering
\includegraphics[clip, trim = 0px 195px 0px 0px,width=0.75\linewidth]{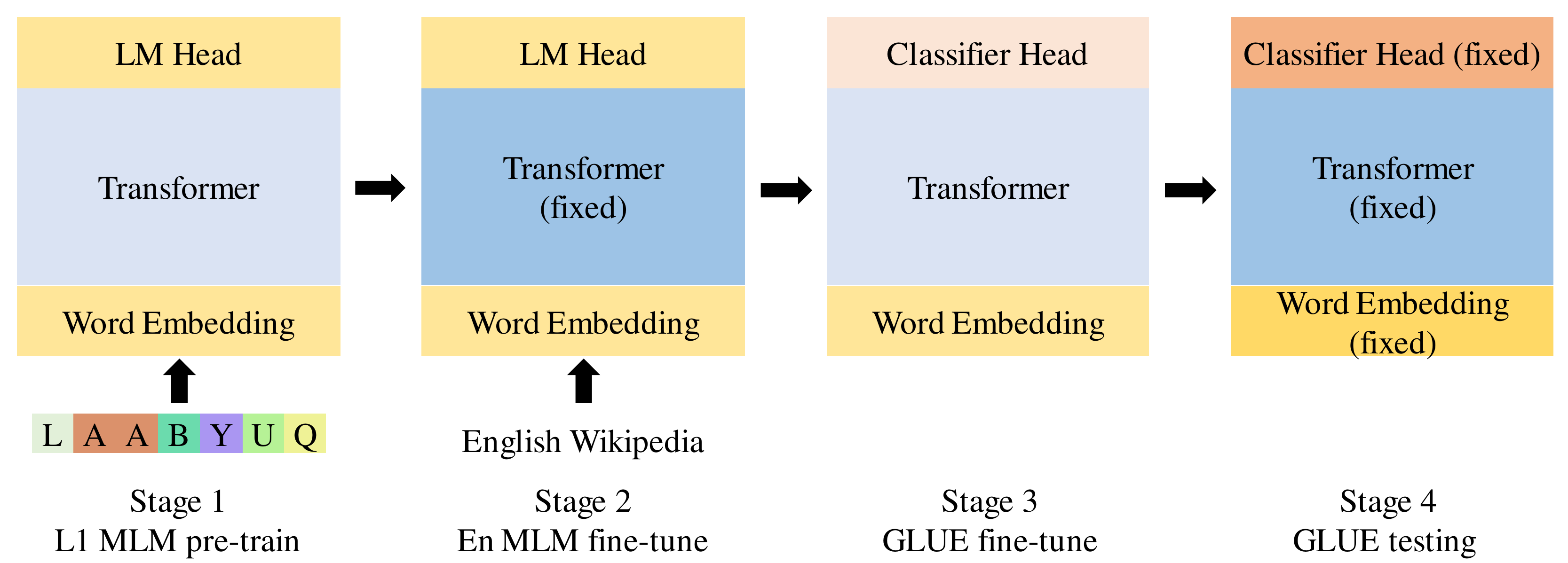}
\caption{Work flow of our experiments for Section~\ref{subsec: MLM finetune}: We first pre-train the whole masked language model on L1 (protein sequence in this figure), and then only fine-tune the word embedding and language model head on English Wikipedia. The third stage is fine-tuning the whole model on English downstream tasks, and the last stage is to test the performance on the fine-tuned downstream task.}
\label{fig:exp finetune}
\end{figure*}

\end{document}